\documentclass{article}

\usepackage{microtype}
\usepackage{graphicx}
\usepackage{subfigure}
\usepackage{booktabs} 

\usepackage{hyperref}

\usepackage[accepted]{icml2024}

\usepackage{amsmath}
\usepackage{amssymb}
\usepackage{mathtools}
\usepackage{amsthm}

\usepackage[capitalize,noabbrev]{cleveref}

\theoremstyle{plain}

\theoremstyle{definition}

\theoremstyle{remark}

\usepackage[textsize=tiny]{todonotes}

\usepackage{amsmath,amsfonts,bm}

\def\1{\bm{1}}

\def\vx{{\bm{x}}}

\DeclareMathAlphabet{\mathsfit}{\encodingdefault}{\sfdefault}{m}{sl}
\SetMathAlphabet{\mathsfit}{bold}{\encodingdefault}{\sfdefault}{bx}{n}

\newcommand{\pdata}{p_{\rm{data}}}

\usepackage{xcolor}
\usepackage{import}
\usepackage{adjustbox}
\usepackage{pifont}

\newcommand{\cmark}{\ding{51}}
\newcommand{\xmark}{\ding{55}}

\newcommand{\norm}[1]{\left\lVert#1\right\rVert}%
\newcommand{\ovx}{\bm{x}^*}

\newcommand{\pcl}{p_{\rm{cl}}}
\newcommand{\pt}{p_{\boldsymbol{\theta}}}

\newcommand{\xzeropred}{\hat{\vx}_0^{(\vx_t)}}
\newcommand{\xzeroprediction}{\hat{\vx}_0^{(\vx_t)}\text{-prediction}}

\newcommand{\decodermeanlearnedshort}{\mu_\theta}

\newcommand{\normaldist}{\mathcal{N}(0, I)}

\def\pt{p_\theta}

\icmltitlerunning{GradCheck}

\begin{document}

\twocolumn[
\icmltitle{GradCheck: Analyzing classifier guidance gradients for conditional diffusion sampling}



\begin{icmlauthorlist}
\icmlauthor{Philipp~Vaeth}{bielefeld,cairo}
\icmlauthor{Alexander~M.~Fruehwald}{cairo}
\icmlauthor{Benjamin~Paassen}{bielefeld}
\icmlauthor{Magda~Gregorova}{cairo}
\end{icmlauthorlist}

\icmlaffiliation{cairo}{Center for Artificial Intelligence, Technical University of Applied Sciences Wuerzburg-Schweinfurt, Franz-Horn-Straße 2,
Wuerzburg, Germany}
\icmlaffiliation{bielefeld}{Bielefeld University, Universitaetsstraße 25, Bielefeld, Germany}

\icmlcorrespondingauthor{Philipp~Vaeth}{philipp.vaeth@thws.de}

\icmlkeywords{Machine Learning, ICML}

\vskip 0.3in
]



\printAffiliationsAndNotice{}  

\begin{abstract}
To sample from an unconditionally trained Denoising Diffusion Probabilistic Model (DDPM), classifier guidance adds conditional information during sampling, but the gradients from classifiers, especially those not trained on noisy images, are often unstable.
This study conducts a gradient analysis comparing robust and non-robust classifiers, as well as multiple gradient stabilization techniques. 
Experimental results demonstrate that these techniques significantly improve the quality of class-conditional samples for non-robust classifiers by providing more stable and informative classifier guidance gradients.
The findings highlight the importance of gradient stability in enhancing the performance of classifier guidance, especially on non-robust classifiers.
\end{abstract}

\section{Introduction}
Denoising Diffusion Probabilistic Models (DDPM)~\cite{ddpm} are state-of-the-art generative models, mapping an intractable data distribution $\pdata$ to a known prior distribution (e.g., $p_T \sim \normaldist$) via a model $\pt(\vx_0) = \int \pt(\vx_0 | \vx_1) \prod_{t=1}^T \pt(\vx_{t-1} | \vx_t) d \vx_{1:T}$. 
Through a Markov chain Gaussian encoder ($\vx_0 \rightarrow \vx_T$) with transitions $q(\vx_t | \vx_{t-1}) :=\mathcal{N}\left(\vx_t ; \sqrt{1-\beta_t} \vx_{t-1}, \beta_t \mathbf{I}\right)$, the data $\vx_0$ is progressively noised with a pre-defined variance schedule $\beta_1, \ldots, \beta_T$.
The Gaussian Markov decoder ($\vx_T \rightarrow \vx_0$) with learned denoising steps $p_\theta\left(\vx_{t-1} \mid \vx_t\right)$ reverses the encoder to produce samples following the data distribution $\pt \approx \pdata$.

An advantage of this type of generative model is the iterative sampling procedure where conditional information can be added without the need for re-training the model, for example through classifier gradients known as \textit{classifier guidance}~\citep{diffusion_thermo,dhariwal2021diffusion}.
For an unconditionally trained DDPM $\pt(\vx_{t-1} \mid \vx_t) = \mathcal{N}(\vx_{t-1}; \mu_t(\vx_t), \Sigma_t(\vx_t))$, the mean $\mu_t$ of the denoising transition can be shifted by the gradients of a classifier trained over the noisy data $\vx_t$ as:
\begin{equation}
    \mu_t' = \mu_t
    + s \, \Sigma_t(\vx_t) \, \nabla_{\vx_t} \log \pcl\left(y \mid \vx_t\right)\enspace ,
    \label{eq:ClassifierGuidance}
\end{equation}
where $s$ is a gradient scaling factor controlling the strength of the classifier guidance.

Classifier guidance is commonly used to add conditional information during inference (e.g., in explainability~\cite{dvce}, in protein design~\cite{gruver2024protein} and in molecular design~\cite{weiss2023guided}).
The main limitation of classifier guidance is that \textit{the classifier needs to be robust to noise similar to that added during the diffusion encoding process}~\cite{dhariwal2021diffusion}.

Training a classifier on the noisy image distribution of the diffusion encoding process makes the classifier robust to the added noise and can be interpreted as a special type of adversarial training~\cite{madry2017towards}.
For cases when training over noisy data is not possible, \citet{avrahami2022blended}~suggested leveraging the diffusion model to get a one-step estimate of the denoised image.
Another challenge besides meaningful gradients due to classifier robustness is the stability of the gradient direction over time.
We therefore apply adaptive moment estimation~\cite{adam} on the classifier gradients as a stabilization technique motivated by similar challenges in stochastic optimization.

Despite these practical solutions to classifier robustness, the intuition why and how the robustness plays a role in the DDPM classifier guidance is still lacking.
In this paper, \textbf{we connect conditional sample quality to classifier robustness and gradient stability. 
We analyze the classifier guidance gradient behavior on non-robust versus robust classifiers, and explore practical solutions for robustness and stability}.

\section{Classifier guidance}\label{sec:classifierguidance}
For class-conditional DDPM samples, equation~\ref{eq:ClassifierGuidance} requires a corresponding scaling factor $s$ trading off diversity for class-consistency, which often requires manual tuning.
This challenge has been addressed by $\ell_2$-normalization of the classifier and DDPM gradients~\cite{dvce}:
\begin{equation}    
    \begin{aligned} 
    \mu_t' &= \mu_t
    +  \frac{s \, \Sigma_t(\vx_t) \, \norm{\mu_t}_2 \, g_t}{\norm{g_t}_2}\enspace , \\
    g_t &= \nabla_{\vx_t} \log \pcl\left(y \mid \vx_t\right) \enspace .
    \end{aligned}
    \label{eq:normalizedclassifierguidance}
\end{equation}
We use this gradient normalization for our experiments, as it keeps the scaling factor $s$ constant across all experimental setups.
\subsection{$\xzeroprediction$}
The major challenge of classifier guidance for non-robust classifiers is the non-informativness of their gradients. 
As the classifier has never seen the DDPM noisy images $\vx_t$ during training, it can not provide meaningful gradients $g_t$ for the guidance.
An intuitive way to directly employ the non-robust classifier during the conditional sampling process is to leverage the DDPM model to obtain a one-step estimate of the denoised image~\cite{avrahami2022blended}: 
\begin{equation}\label{eq:xzeropred}
    \xzeropred=\frac{\vx_t}{\sqrt{\bar{\alpha}_t}}-\frac{\sqrt{1-\bar{\alpha}_t} \epsilon_\theta\left(\vx_t, t\right)}{\sqrt{\bar{\alpha}_t}}\enspace ,
\end{equation}
with $\bar{\alpha}_t=\prod_{s=1}^t\left(1-\beta_s\right)$.
We explore this modification in our experiments by setting $g_t = \nabla_{\vx_t}\big[\log \pcl\left(y \mid \xzeropred\right)\big]$ for equation~\ref{eq:normalizedclassifierguidance} as a potential solution for obtaining meaningful gradients earlier in the guided sampling process.
The main limitation of this modification is that the gradient needs to be calculated across the one-step denoising process for each sampling step, greatly increasing memory requirements and sampling time.

\subsection{Adaptive moment estimation (ADAM)}
An intuitive way to interpret the classifier guidance is as a multi-objective stochastic optimization problem, where $\decodermeanlearnedshort$ is the implicit gradient of the DDPM model towards high probability w.r.t the data distribution, and $g_t$ is the gradient towards maximum classifier confidence w.r.t the target class.
This interpretation allows us to employ techniques of gradient stabilization well-explored in the field of stochastic optimization.
Concretely, we introduce ADAM~\citep{adam} into the guidance steps to automatically adapt the learning rates from estimates of the first and second moments of the gradients:
\begin{equation}
    g_t = \nu(\nabla_{\vx_t} \log \pcl\left(y \mid \vx_t\right))\enspace ,
    \label{eq:ADAM}
\end{equation}
where $\nu()$ indicates the ADAM adjustment of the gradient. 
In equation~\ref{eq:ADAM}, $\vx_t$ can also be replaced by $\xzeropred$.
We use ADAM with default parameters for stochastic optimization ($\beta_1=0.9, \beta_2=0.999$) and without a dedicated step size ($\eta=1$).
Other algorithms or hyper-parameter settings may be explored in future work.

\section{Experimental setup}
In the following section, we will first introduce the experimental setup.
We then explain how we analyze the gradients and which metrics we use to verify the results.
We introduce a custom data set for our experiments and train a DDPM model, as well as two classifiers on this data set.
The code for all experiments and Weights\&Biases experiment tracking is provided for training and sampling~(\url{https://anonymous.4open.science/r/gradcheck}).
All training and sampling runs were conducted on a single Nvidia A100 80 GB GPU.
More details of the experimental setup can be found in the appendix.

\subsection{Data set}
Common image classification dataset such as ImageNet~\cite{imagenet} are often too complex with unclear class boundaries (e.g., no clear class assignments, multiple concepts in one image, overlapping concepts between classes), which makes showing the effectiveness of gradient stabilization techniques ambiguous.
We therefore create a synthetic data set called SportBalls with clear classes and concepts.

The data set is generated by randomly selecting one out of four sport balls and placing it at random coordinates on white background with random rotation and scaling. 
The data set is carefully created to have similar objects (i.e., scaling, shape, size, rotation and placement) but with clear semantic differences (i.e., colors and pattern).
\subsection{DDPM}
We train a standard DDPM~\cite{ddpm,dhariwal2021diffusion} based on the state-of-the-art implementation framework Diffusers~\cite{von-platen-etal-2022-diffusers}.
The 22.5M parameter model is trained for 1000 epochs (almost 3 days).
Unconditional generations can be found in the appendix.

\subsection{Classifiers} \label{sec:classifiers}
We train one robust and one non-robust MobileNet~\cite{howard2019searching} classifier. 
The training script and the classifier architecture is the same for both the robust and the non-robust classifier with the only difference being the data on which the classifiers are trained and evaluated.

The non-robust classifier is trained and evaluated on clean images and achieves a validation accuracy of 99.2\%.
Not being trained on any noisy images, the non-robust classifier achieves a validation accuracy of 26.65\% on the noisy DDPM latents.
The robust classifier is trained on the noisy latents ($\ovx_1, \dots,\ovx_T$) based on uniform sampling of time steps and the noise schedule of the DDPM model.
This means the classifier is fully aware of the noisy data distribution of the guided sampling process and should have meaningful gradients.
The robust classifier achieves a validation accuracy on the noisy latents of 64.1\% ($>$99\% on the clean images), which is a good performance considering the third most noised latents have no perceptible structure. 

\subsection{Conditional sampling}
The conditional samples are generated based on the definitions in section~\ref{sec:classifierguidance}.
All experiments are conducted with constant hyper-parameters (i.e., a step size of $s=0.04$ and a batch size of 64).
The step size is based on a preliminary experiment to find the optimal trade-off between sample quality and the amount of class-conditioning for the robust guiding classifier without modifications (see appendix).
The batch size was chosen due to GPU memory constraints, while still providing a large enough sample size to calculate average and standard deviation for the experiments.

\subsection{Gradient analysis}
To monitor the classifier gradients over the diffusion sampling process, we calculate the cosine similarity of each time step compared to the previous time step:
\begin{equation}
    \cos(\vx_t)=\frac{c_t\cdot c_{t-1}}{\norm{c_t}_2\,\norm{c_{t-1}}_2}\enspace ,
\end{equation} 
where $c_t$ is the conditioning term (right side of equation~\ref{eq:normalizedclassifierguidance}, i.e., $\mu_t'-\mu_t$).
This score over time tracks the gradient directions and shows the efficiency of the stabilization techniques. 
We show this score in the experiments as the mean and standard deviation over a batch of 64 samples to account for the randomness in the DDPM sampling process.

For interpretation, we note two properties of the cosine similarity in high-dimensional spaces: 
First, even values close to one permit large differences along a few dimensions. 
Second, if vectors are randomly drawn, their cosine similarity is likely close to zero (i.e., they are almost orthogonal).

\subsection{Metrics}
We evaluate the resulting conditional samples based on classifier accuracy and FID~\cite{fid}, as well as providing the samples as images in the appendix.
The classifier accuracy is based on the same classifier used in the guidance process and verifies that the resulting samples are indeed classified as the target class by the guiding classifier.
The FID compares distributions of images and is a common measure for visual quality of the images.
For our use-case of generating conditional samples, we will compare the distribution of 64 training images from the target class to 64 generated conditional samples.
The class-conditional FID measure in combination with the classification accuracy indicates how successful the guidance was in generating high-quality class-conditional samples.

\section{Experimental results \& discussion}
\begin{figure}[h!]
    \centering
    \includegraphics[width=0.9\linewidth]{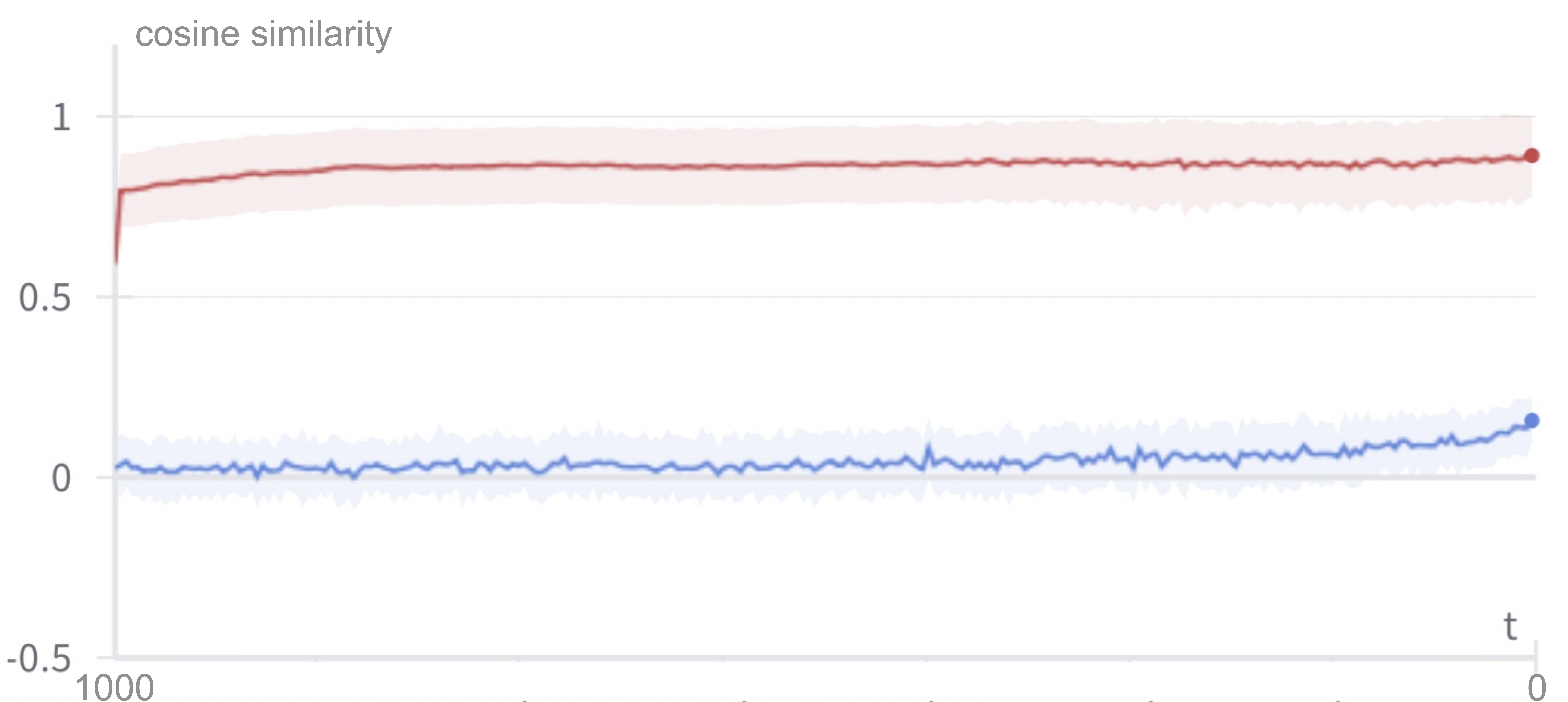}
    \caption{Classifier guidance cosine similarity between time steps for the robust classifier (red) and the non-robust classifier (blue). The values are presented as the mean over a batch of 64 generations, with the lighter colors indicating the standard deviation.}
    \label{fig:robust_nonrobust}
\end{figure}

\begin{figure}[h!]
    \centering
    \includegraphics[width=0.88\linewidth]{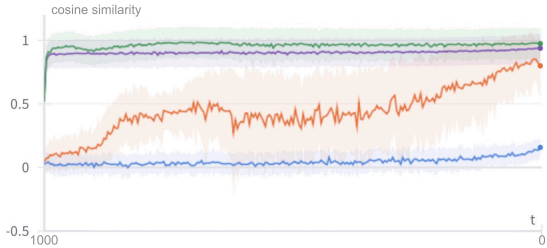}
    \caption{Classifier guidance without modifications (blue), with only $\xzeroprediction$ (orange), with only ADAM stabilization (purple), and with both modifications (green) cosine similarity between time steps for the non-robust classifier. The values are presented as the mean over a batch of 64 generations, with the lighter colors indicating the standard deviation.} 
    \label{fig:nonrobustall}
\end{figure}

As a baseline, we run the classifier guidance with equation~\ref{eq:normalizedclassifierguidance} for a robust and a non-robust classifier (see section~\ref{sec:classifiers}). 
In figure~\ref{fig:robust_nonrobust}, we can see the robust classifier (red) having a high and fairly consistent cosine similarity over time, whereas the non-robust classifier (blue) has almost orthogonal conditioning vectors between time steps.
Based on the intuition that random vectors in high-dimensional spaces are likely to be orthogonal, we can infer that the non-robust classifier has almost random gradients for most of the denoising process.
This difference in gradient stability is verified by the metrics, where the robust classifier guidance has over 80\% points more accuracy and 100 points lower FID score (see table~\ref{tab:metrics}).
\begin{table}[H]
    \begin{adjustbox}{width=0.95\columnwidth,center}
    \begin{tabular}{@{}l|cccc|cccc@{}}
        \toprule
        Classifier        & \multicolumn{4}{c}{Robust Classifier} \vline& \multicolumn{4}{c}{Non-robust Classifier} \\\midrule
        $\xzeroprediction$     & \xmark  &  \cmark &  \xmark & \cmark & \xmark  & \cmark & \xmark & \cmark  \\ 
    ADAM                     &  \xmark &  \xmark &   \cmark &   \cmark & \xmark  & \xmark  & \cmark & \cmark \\ \midrule
        FID $\downarrow$ & 26.641 & 38.053 & \textbf{26.575}   & 34.484 & 126.526 & 103.301 & 127.526 &  \textbf{24.348} \\
        Acc $\uparrow$ & \textbf{96.88\%}  & 81.25\% & 92.19\% & 78.13\% & 15.63\%  & 39.06\% & 15.63\% &  \textbf{85.94\%} \\
        \bottomrule
    \end{tabular}
    \end{adjustbox}
    \caption{Metric table for all experimental setups.} 
    \label{tab:metrics}
\end{table}
The non-robust model yields unsatisfactory class-conditional samples, as it has not been trained on noisy images (see appendix).
To get around this limitation, we employ the $\xzeroprediction$ (equation~\ref{eq:xzeropred}) to get more informative and more stable gradients.
We can see in figure~\ref{fig:nonrobustall} (orange line) that the $\xzeroprediction$ indeed stabilizes the gradients over time with a much higher cosine similarity than without the modification.
Especially towards the end of the sampling process, the cosine similarity increases to similar numbers as that of the robust classifier in figure~\ref{fig:robust_nonrobust}.
The metric in table~\ref{tab:metrics} support these findings with a decrease in FID of around 23 points and an increase in accuracy of over 23\% points.

The non-robust classifier with $\xzeroprediction$ has more informative gradients sooner in the sampling process, and the stability increases over time, which is consistent with the intuition that the one-step denoising estimate of the DDPM should get better the further the denoising process is.
\newpage
To further stabilize the gradients, we apply ADAM on the gradients, which computes an adaptive step size based on first and second moments of the gradients.
The resulting cosine similarity over time (green line in figure~\ref{fig:nonrobustall}) very quickly reaches almost one which indicates the conditioning \textit{picks one direction} towards maximum classifier confidence w.r.t the target class and sticks to this direction with a very high cosine similarity.
Both the $\xzeroprediction$ and the ADAM gradient stabilization combined increase the accuracy of the final samples to almost 86\% and decreases the FID to around 24, which is even lower than the FID of the robust classifier without any modification (see table~\ref{tab:metrics}).

Applying only the ADAM gradient stabilization to the non-robust classifier guidance seems to improve the gradient stability (purple line in figure~\ref{fig:nonrobustall}), but results in almost unchanged metrics compared to no modifications (see table~\ref{tab:metrics}).
This underlines two key requirements for high quality class-conditional samples through classifier guidance: The gradients have to be meaningful and stable over time.

\textbf{The experiments indicate that both the $\xzeroprediction$ and the ADAM stabilization contribute to higher quality class-conditional samples for a non-robust guiding classifier by providing more meaningful gradients earlier in the denoising process and by keeping the gradient directions more consistent over time.}

\begin{figure}[h!]
    \centering
    \includegraphics[width=0.88\linewidth]{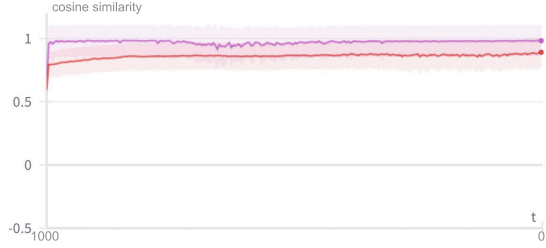}
    \caption{Robust classifier guidance cosine similarity between time steps with ADAM gradient stabilization and $\xzeroprediction$ (purple) and without stabilization (red). The values are presented as the mean and standard deviation over a batch of 64 generations.} 
    \label{fig:robustxzeropredadam}
\end{figure}

Lastly, we verify how the gradient stabilization techniques behave on the already robust guiding classifier.
We employ the $\xzeroprediction$ and the ADAM gradient stabilization, and observe an even higher classifier gradient consistency than the regular robust classifier (see figure~\ref{fig:robustxzeropredadam}).
However, the metrics in table~\ref{tab:metrics} decrease (-19\% points accuracy and +7.8 FID) and the samples are visually less class-consistent with those of the target-class.
We assume that this is due to over-regularization of the gradients, which has a negative effect on already meaningful gradients.
\section{Conclusion}
In this paper, we conducted an analysis of classifier guidance gradients in Denoising Diffusion Probabilistic Models (DDPMs) and explored methods to enhance the stability and effectiveness of these gradients. 
Our study highlights the challenges posed by non-robust classifiers, which are not trained on noisy data, leading to unstable and less informative gradients during the conditional sampling process.

To address these challenges, we analyze two key techniques: $\xzeroprediction$ and Adaptive Moment Estimation (ADAM). 
The $\xzeroprediction$ method leverages the DDPM model to provide a one-step clean estimate of the noisy latents, resulting in more stable and informative gradients. 
ADAM, on the other hand, adapts the learning rates based on the first and second moments of the gradients, further stabilizing the gradient directions over time.

Our experimental results show that these techniques significantly improve the quality of class-conditional samples generated by non-robust classifiers. 
The $\xzeroprediction$ method alone increased the cosine similarity of gradients and improved the classification accuracy and FID scores. 
When combined with ADAM, the improvements were even more pronounced, achieving higher accuracy and lower FID scores than the robust classifier without modifications.

Interestingly, applying the same stabilization techniques to an already robust classifier resulted in over-regularization, leading to a decrease in sample quality. 
This suggests that while gradient stabilization is crucial for non-robust classifiers, it must be carefully tuned for robust classifiers to avoid negative effects.

In conclusion, our findings underscore the importance of meaningful and stable classifier guidance gradients for generating high-quality class-conditional samples from an unconditional DDPM. 
The proposed methods provide practical solutions for a more robust and effective classifier guidance. 
Future work should explore optimal hyper-parameter settings and other stochastic optimization algorithms.

\clearpage
\bibliography{main}
\bibliographystyle{icml2024}

\newpage
\appendix
\onecolumn
\section{Data set}
The SportBalls data set (3x64x64) is synthetically created by randomly selecting one out of four sport balls (serving as the classes) and placing it at random coordinates on white background with random rotation and random scaling.
The data set is created to have similar objects (i.e., scaling, shape, size, rotation and placement) but with clear semantic differences (i.e., the color and the pattern of the sport balls).
The images used in the data set are by Freepik (\url{https://www.flaticon.com/authors/freepik}).
The final data set images can be seen in figure~\ref{fig:gt_images}.
\section{DDPM}
\begin{figure}[h!]
    \hfill
    \subfigure[Ground-truth images\label{fig:gt_images}]{\includegraphics[width=7.5cm]{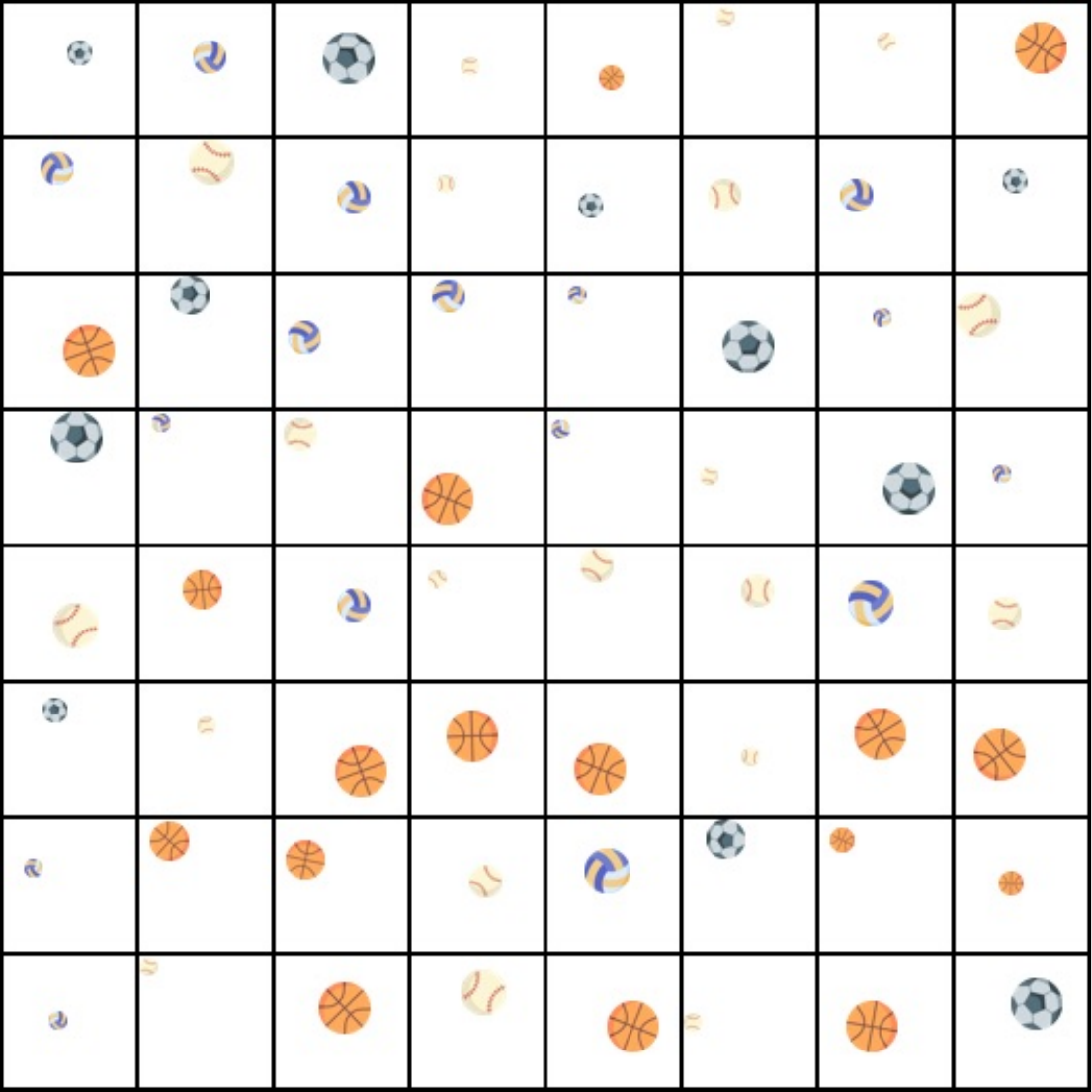}}
    \hfill
    \subfigure[Unconditional DDPM samples\label{fig:samples_unconditional}]{\includegraphics[width=7.5cm]{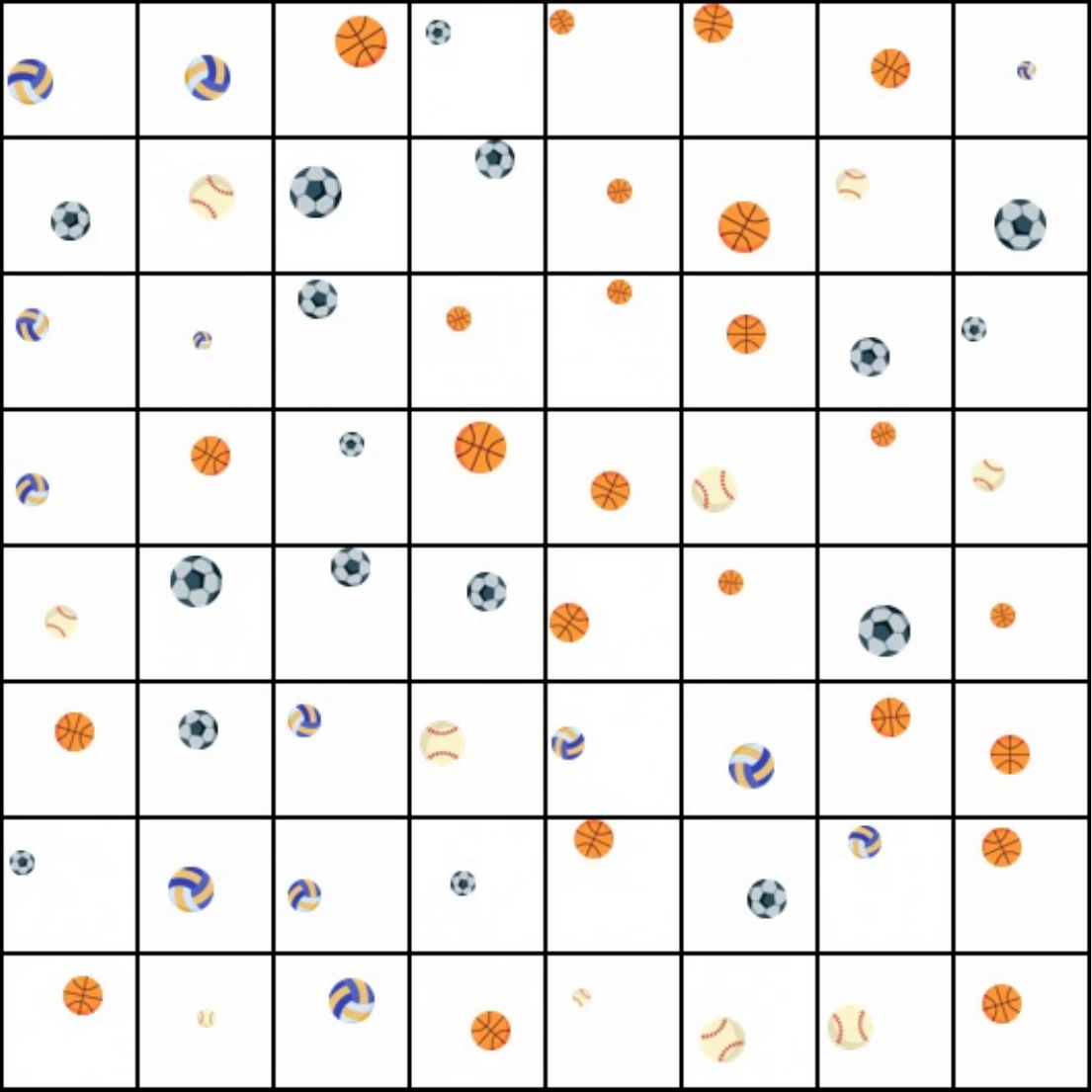}}
    \hfill
    \caption{SportBalls images}
\end{figure}

The DDPM was trained using the state-of-the-art implementation framework Diffusers~\cite{von-platen-etal-2022-diffusers} with a linear noise schedule of $\beta_1=0.0001, \beta_T=0.02$.
The model was trained unconditionally for $T=1000$ time steps with a fixed variance and through the simplified noise prediction loss~\cite{ddpm}.
The underlying U-Net noise predictor has about 22.5 million parameters and took almost 3 days to train on a single NVIDIA A100 80 GB GPU.
The resulting unconditional samples are shown in figure~\ref{fig:samples_unconditional}.

\section{Classifier Guidance Ablation Study}
In this section, we explore different hyper-parameter settings and supplement the experiments from the main paper.
All classifiers used in this paper are MobileNet~\cite{howard2019searching} ``version\_3\_small'' models with 1.5M parameters.
The models were trained for 200 epochs with a standard training procedure and an early stopping criterion on the validation accuracy ($\geq$99\%).
The final training time was around 16 minutes for the robust classifier and 2 minutes for the non-robust classifier.
The target class for all experiments is the class ``baseball''. 

\subsection{Classifier guidance scale}
\begin{figure}[H]
    \centering
    \includegraphics[width=1\linewidth]{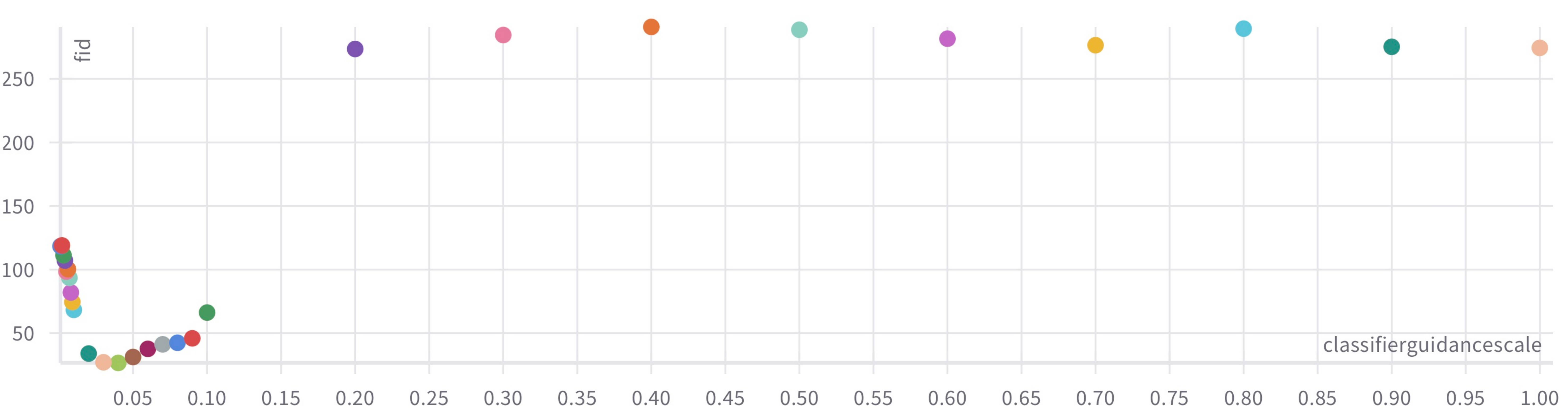}
    \caption{Trade-off between the probability w.r.t. the target class image distribution (FID) and the classifier guidance scale ($s$)}
    \label{fid:fid_vs_guidance}
\end{figure}
The best FID to classifier guidance scale (s) trade-off was achieved at $s=0.04$ and was therefore used in the experiments shown in the main paper.

\subsection{Supplementary images for the experiments}
In this section, we provide the conditionally sampled images corresponding to the experiments conducted in the main paper.

\begin{figure}[h!]
    \centering
    \includegraphics[width=1\linewidth]{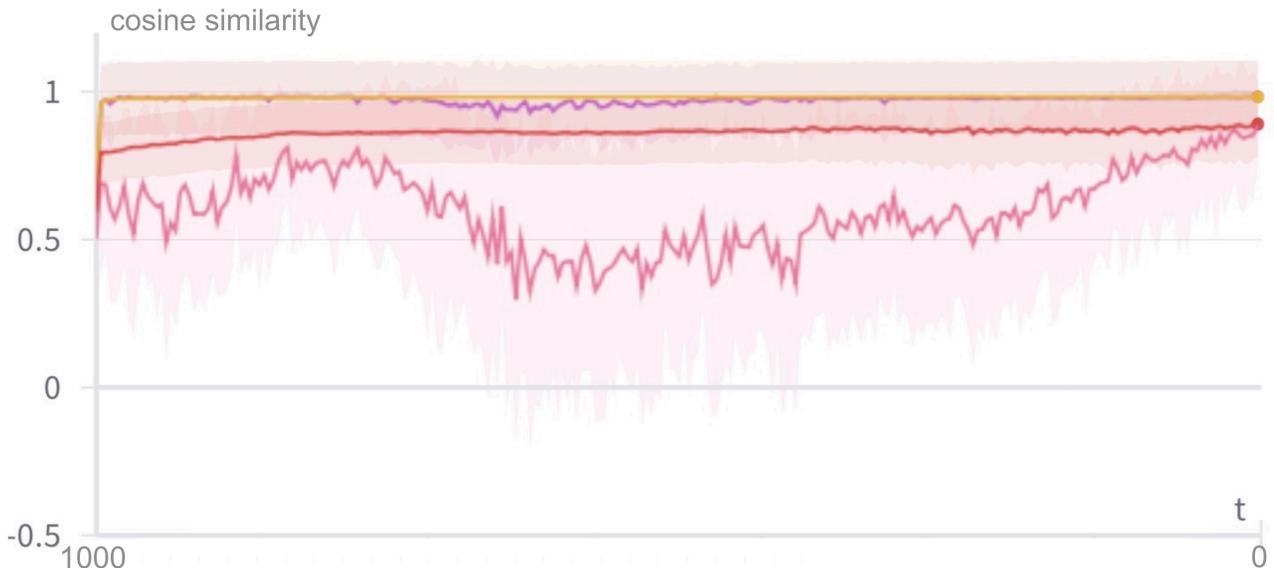}
    \caption{Classifier guidance without modifications (red), with only $\xzeroprediction$ (pink), with only ADAM stabilization (yellow), and with both modifications (magenta) cosine similarity between time steps for the non-robust classifier. The values are presented as the mean over a batch of 64 generations, with the lighter colors indicating the standard deviation.} 
    \label{fig:robustall}
\end{figure}

Figure~\ref{fig:robustall} shows the cosine similarity over time for the robust classifier and all combinations of modifications. 

\begin{figure}[h!]
    \centering
    \subfigure[ADAM + $\xzeroprediction$]{\includegraphics[width=7.5cm]{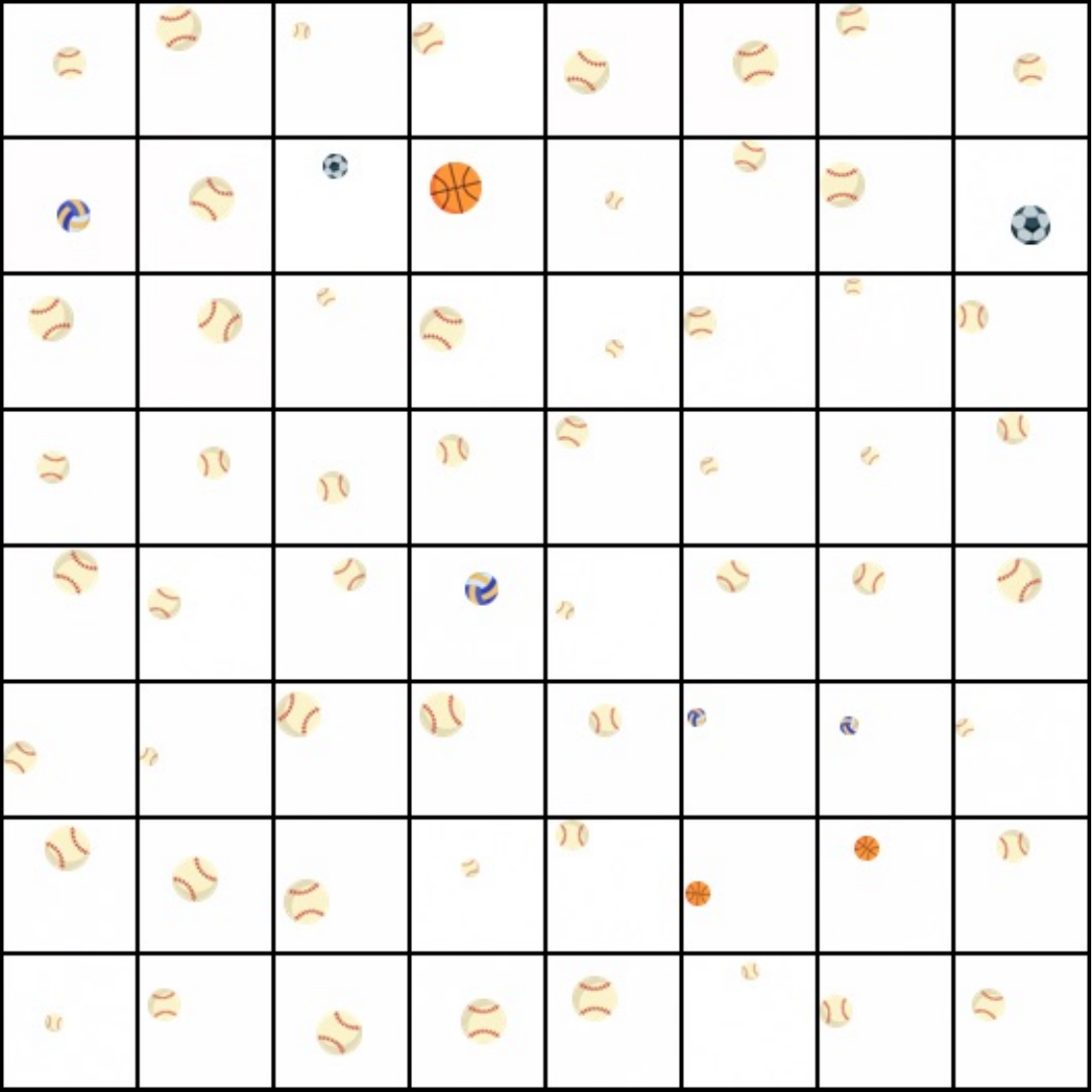}}
    \subfigure[ADAM]{\includegraphics[width=7.5cm]{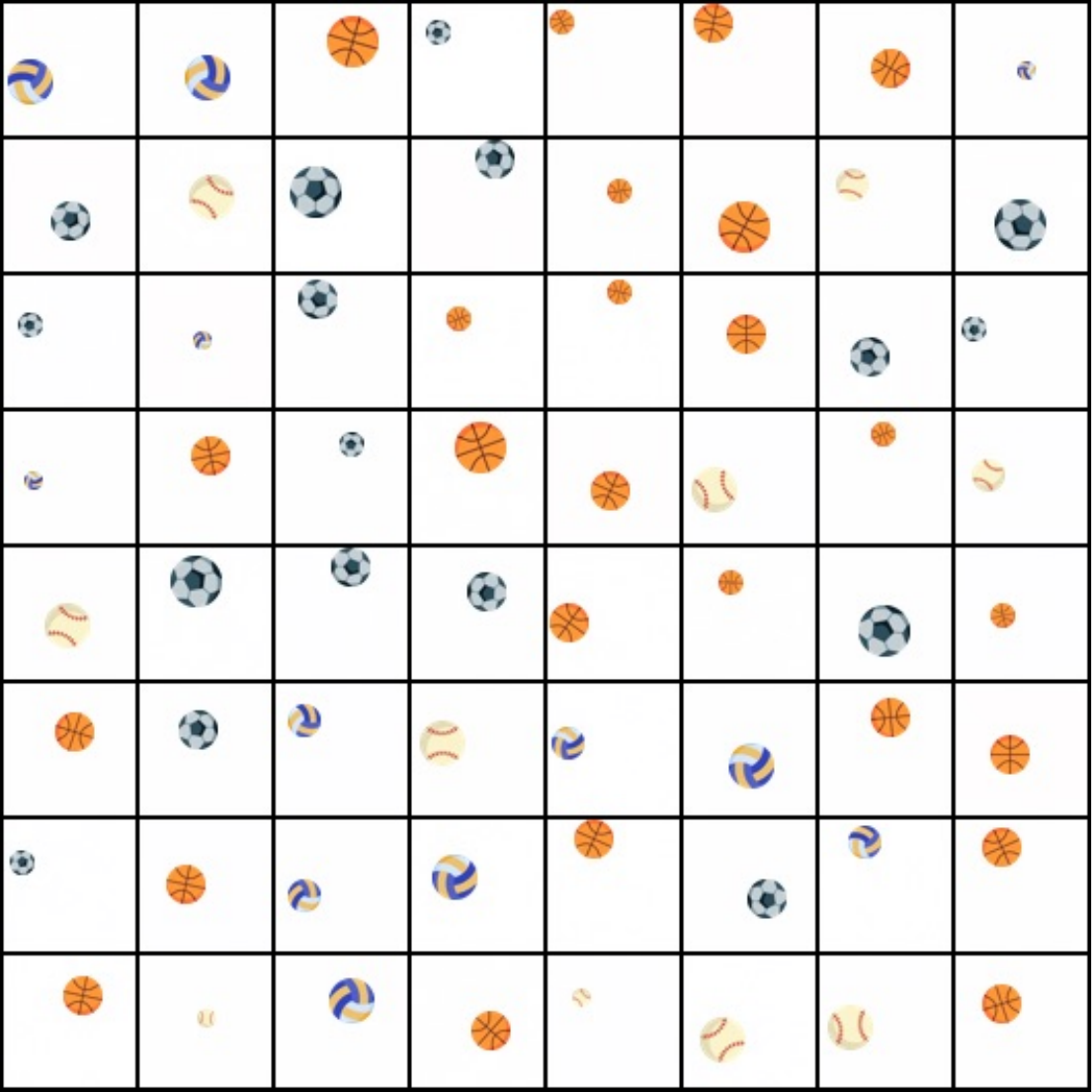}}\\
    \subfigure[$\xzeroprediction$]{\includegraphics[width=7.5cm]{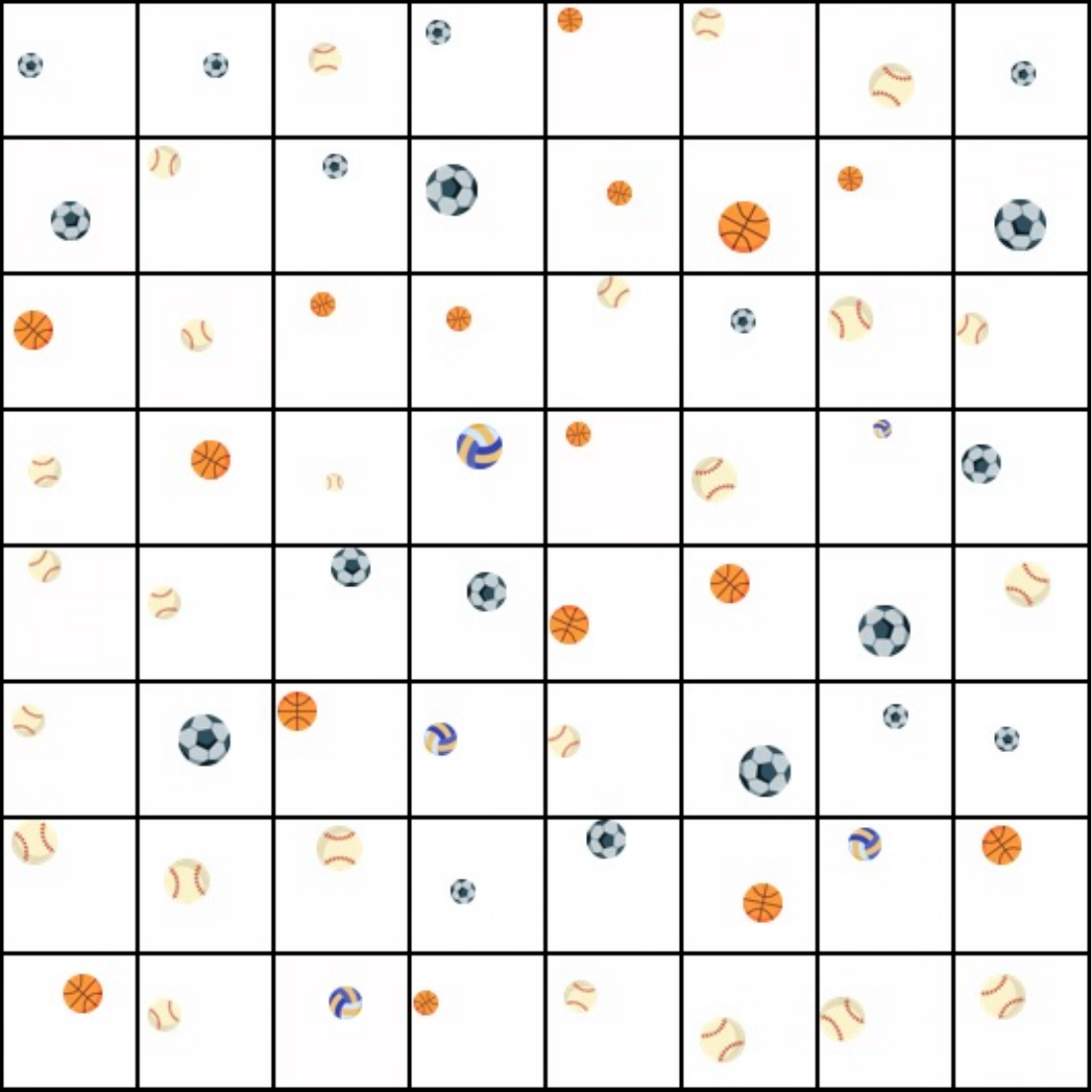}}
    \subfigure[No stabilization]{\includegraphics[width=7.5cm]{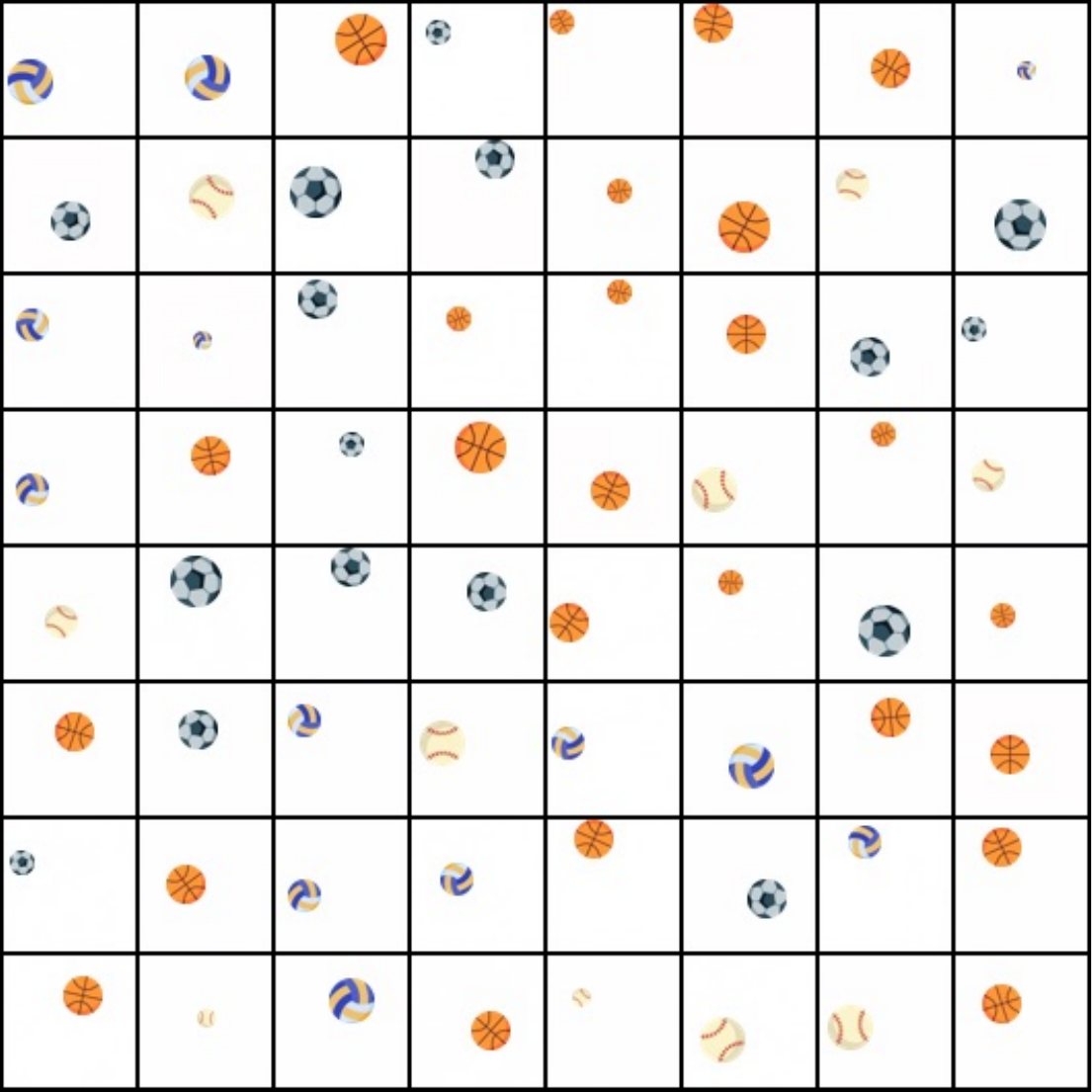}}
    \caption{Supplementary samples for the non-robust classifier guidance with and without $\xzeroprediction$ + ADAM gradient stabilization}
    \label{fig:nonrobust_samples_all}
\end{figure}

\begin{figure}[h!]
    \centering
    \subfigure[ADAM + $\xzeroprediction$]{\includegraphics[width=7.5cm]{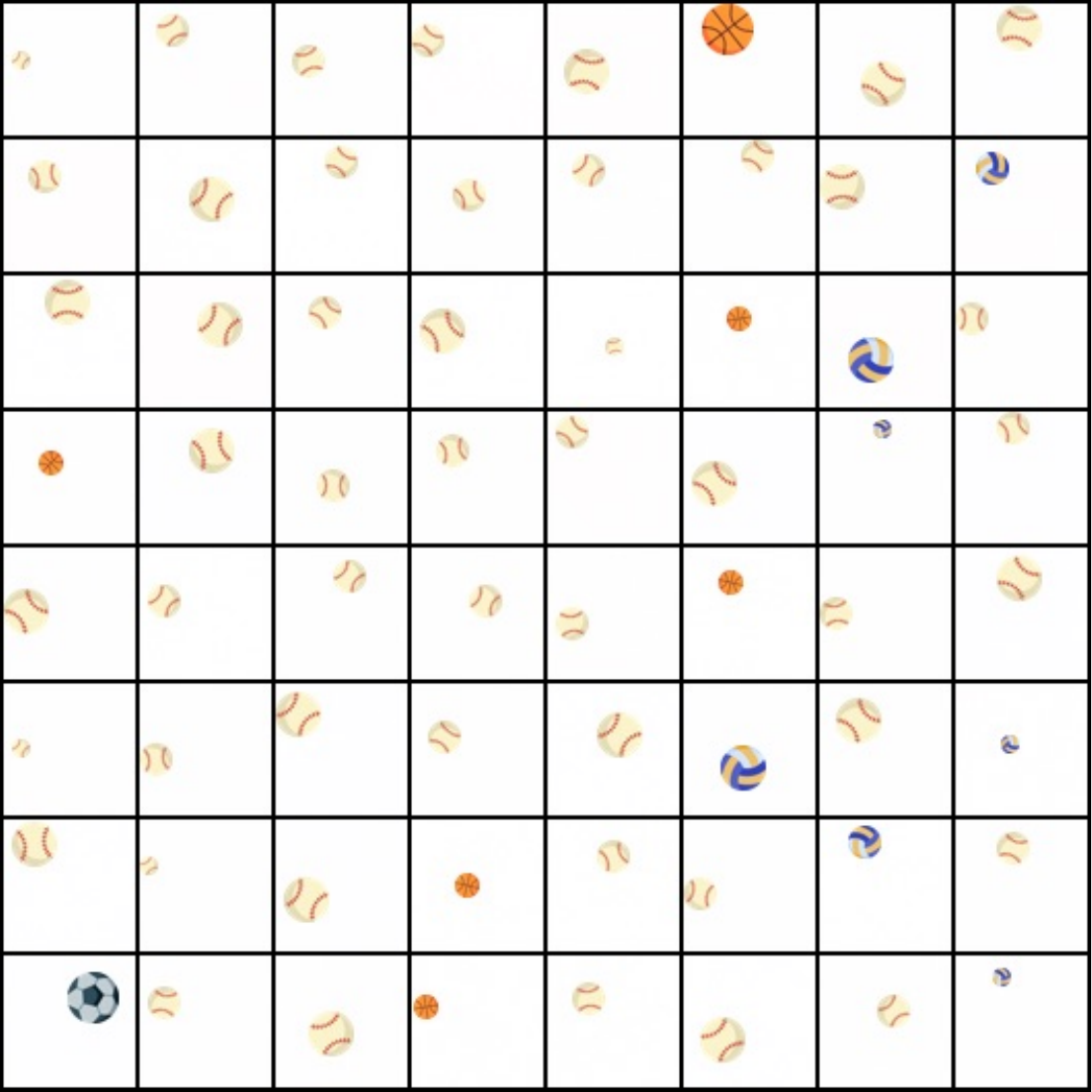}}
    \subfigure[ADAM]{\includegraphics[width=7.5cm]{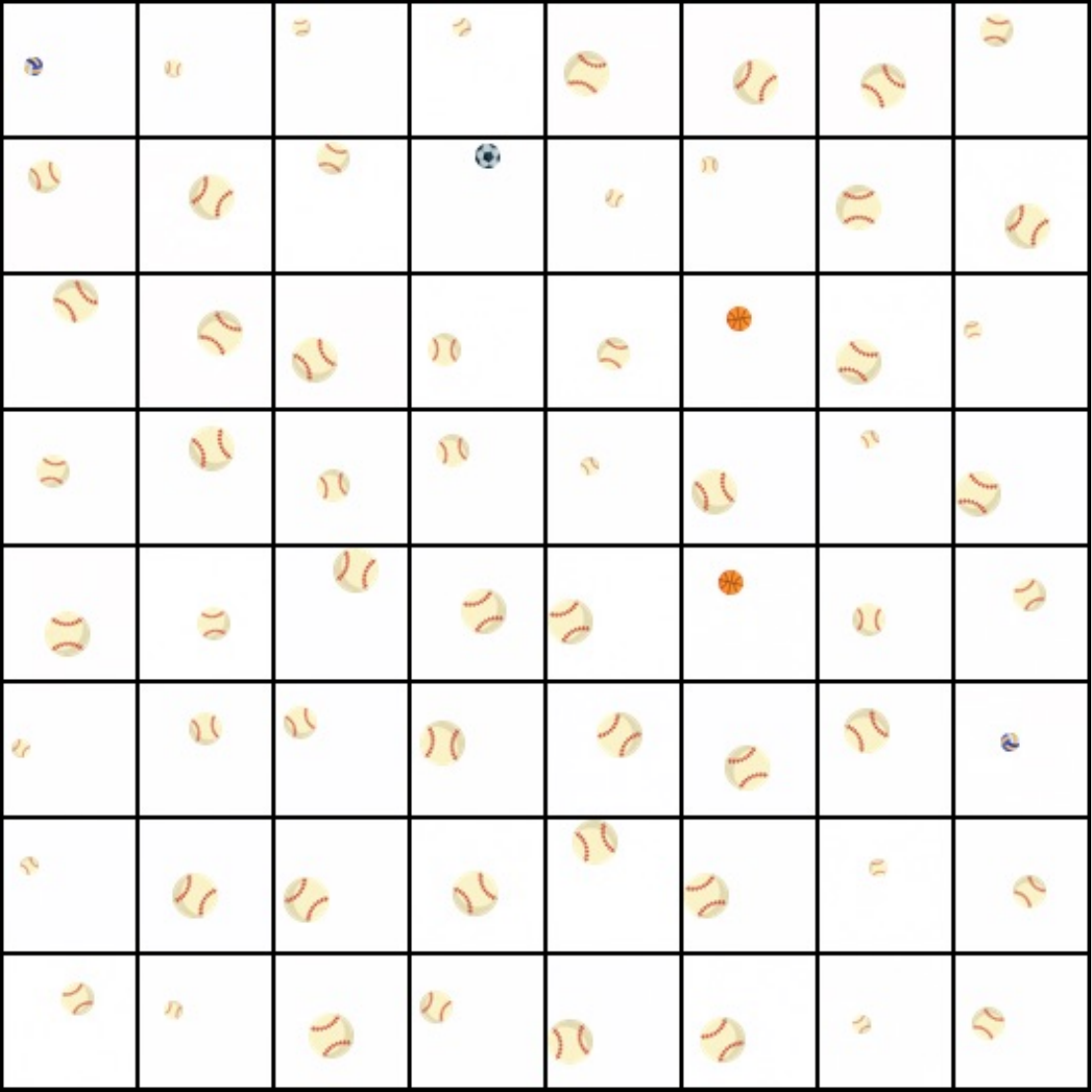}}\\
    \subfigure[$\xzeroprediction$]{\includegraphics[width=7.5cm]{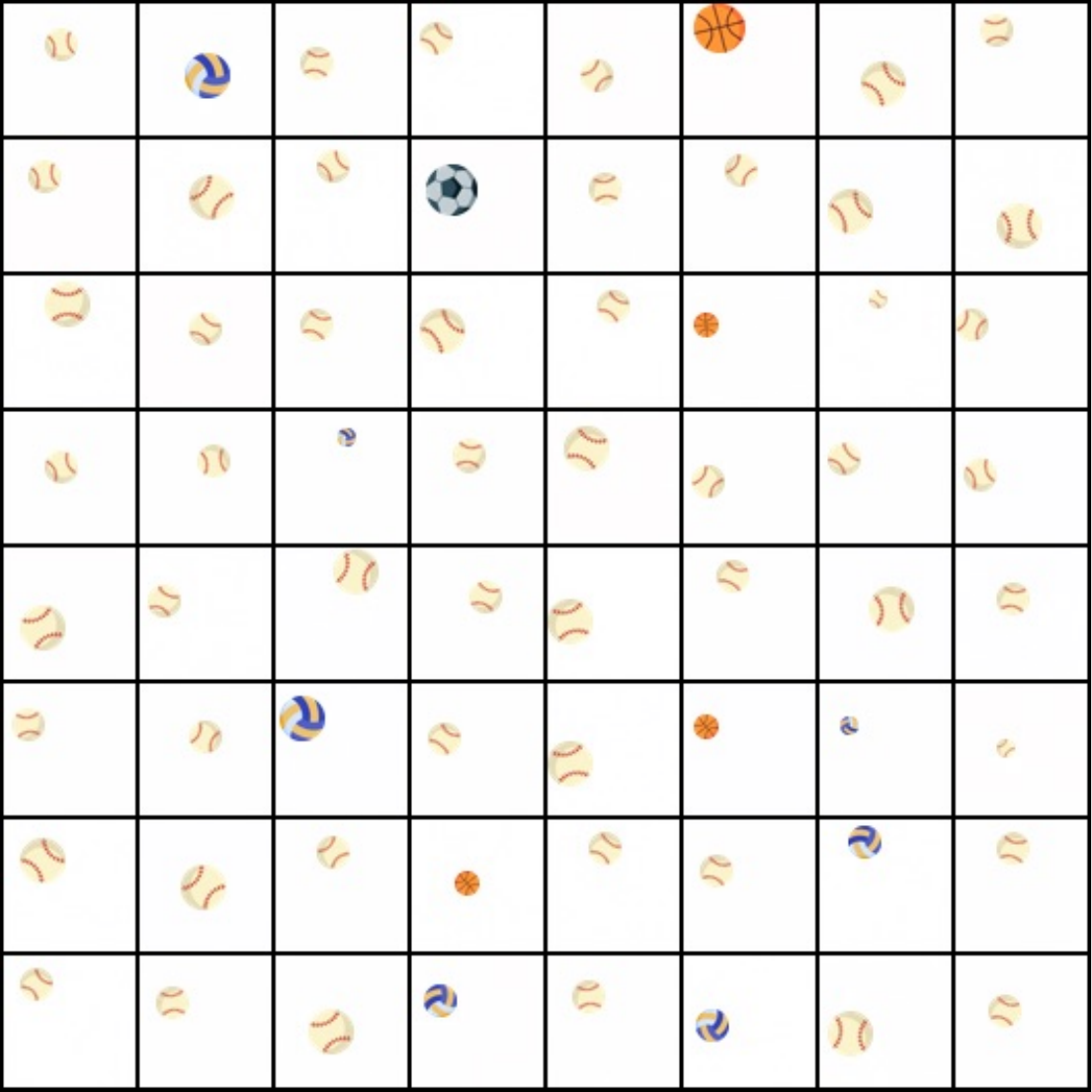}}
    \subfigure[No stabilization]{\includegraphics[width=7.5cm]{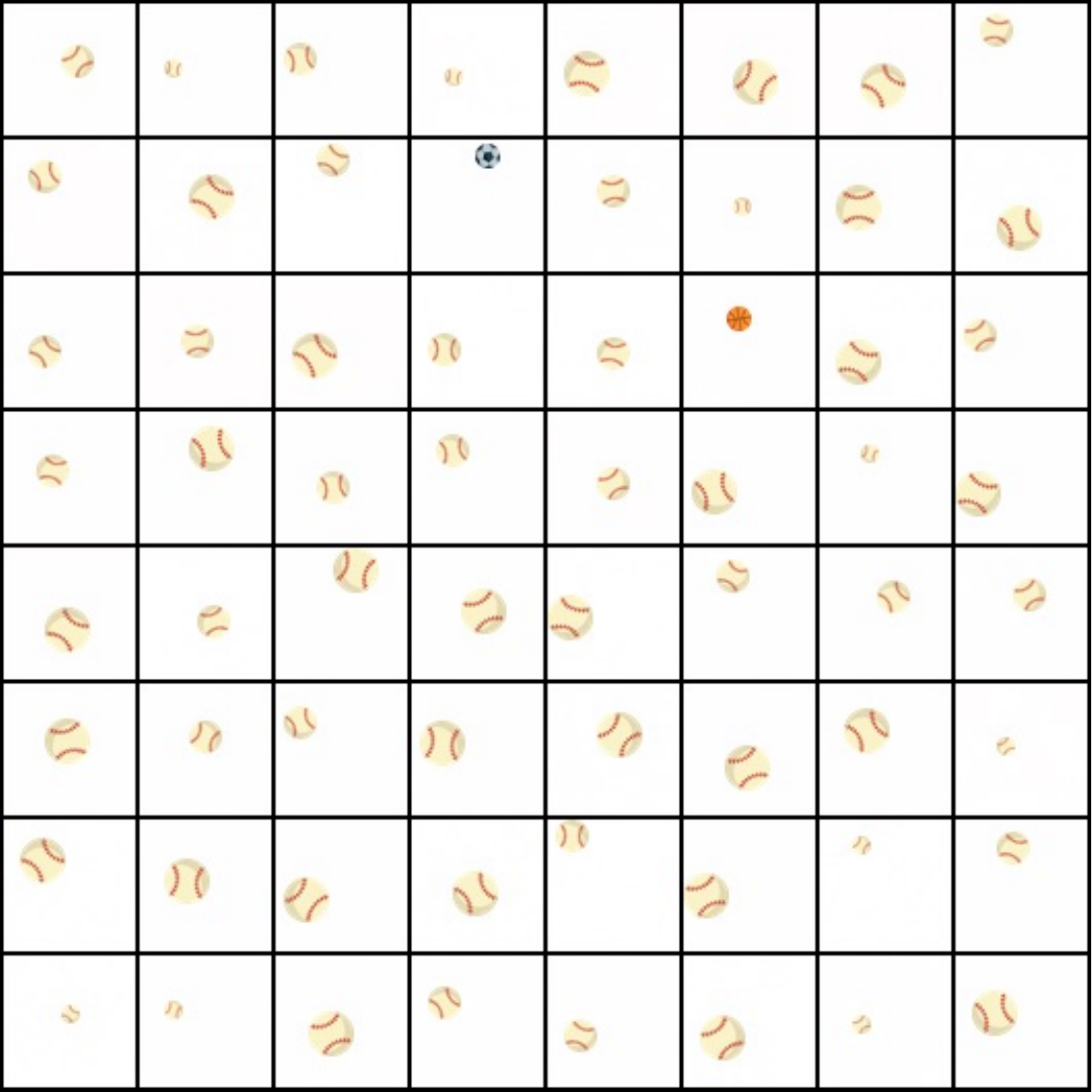}}
    \caption{Supplementary samples for the robust classifier guidance with and without $\xzeroprediction$ + ADAM gradient stabilization}
    \label{fig:robust_xzeropred_adam_samples}
\end{figure}

\vspace{5cm}
Figure~\ref{fig:nonrobust_samples_all} shows the samples for the \textbf{non-robust} classifier with the $\xzeroprediction$ and ADAM stabilization and their combinations.
\newpage
Figure~\ref{fig:robust_xzeropred_adam_samples} shows the samples for the \textbf{robust} classifier with the $\xzeroprediction$ and ADAM stabilization and their combinations.

\end{document}